# Identifying Player's Strategies in No Limit Texas Hold'em Poker through the Analysis of Individual Moves

Luís Filipe Teófilo and Luís Paulo Reis

Departamento de Engenharia Informática, Faculdade de Engenharia da Universidade do Porto, Portugal

Laboratório de Inteligência Artificial e de Ciência de Computadores, Universidade do Porto, Portugal

luis.teofilo@fe.up.pt, lpreis@fe.up.pt

**Abstract.** The development of competitive artificial Poker playing agents has proven to be a challenge, because agents must deal with unreliable information and deception which make it essential to model the opponents in order to achieve good results. This paper presents a methodology to develop opponent modeling techniques for Poker agents. The approach is based on applying clustering algorithms to a Poker game database in order to identify player types based on their actions. First, common game moves were identified by clustering all players' moves. Then, player types were defined by calculating the frequency with which the players perform each type of movement. With the given dataset, 7 different types of players were identified with each one having at least one tactic that characterizes him. The identification of player types may improve the overall performance of Poker agents, because it helps the agents to predict the opponent's moves, by associating each opponent to a distinct cluster.

## 1 Introduction

Poker is a game that is became a field of interest for the AI research community on the last decade. This game presents a radically different challenge when compared to other games like chess or checkers. In these games, the two players are always aware of the full state of the game. This means that it is possible to know the opponent strategy just by observing the movement of the game pieces. Unlike that, Poker game state is hidden because each player can only see its cards or community cards, and therefore it's much more difficult to detect the opponents' strategies. Poker is also stochastic game i.e. it admits the element of chance.

The announced characteristics of Poker make it essential to model the opponents before making a decision [1,2]. By identifying the opponents playing style, it is possible to predict their possible actions and therefore make a decision that has better probability of success.

This article focuses on the identification of new playing styles, through the analysis of game moves. The analysis of game moves was done with clustering algorithms.

Thus, the purpose of this study is to identify groups of similar game moves. By identifying these move groups, we can characterize an opponent by knowing the group of moves that it uses most.

The goals of this work are:
- Process a significant amount of poker game logs between human players, from casino clients;
- Extract information about the events of the games presented on the game logs;
- Use clustering algorithms to identify groups of common game moves;
- Measure the players' relative frequency of use of the identified move groups in order to identify common strategies.

The rest of the paper is organized as follows. Section 2 describes the game of Poker and the variant of Poker that was studied – No Limit Texas Hold'em. Section 3 describes related work about some approaches followed earlier to create Poker agents as well as the description of the type of clustering algorithms that were used in this work. Section 4 presents a general overview of this work, including the steps of development from the extraction of Poker hands to the definition of strategies. Section 5 presents the steps that were followed to extract information from Poker games, present on the game logs. Section 6 presents the usage of clustering algorithms to determine types of actions in Poker games. Section 7 presents the usage of clustering algorithms to determine types of strategies, based on the frequency of each type of action. Finally, section 8 presents the paper main conclusions and some pointers for future work.

## 2   Texas Hold'em Poker

Poker is a generic name for literally hundreds of games, but they all fall within a few interrelated types [3]. It is a card game in which players bet that their hand is stronger than the hands of their opponents. All bets go into the pot and at the end of the game the player with the best hand wins. Another way of winning is making the other players forfeit the hand and therefore being the last standing player.

### 2.1  No Limit Texas Hold'em

No Limit Texas Hold'em is a Poker variant that uses community cards. At the beginning of every game, two cards are dealt to each player. A dealer player is assigned and marked with a dealer button. The dealer position rotates clockwise from game to game. After that, the two players to the left of dealer post the blind bets (small-blind and big-blind). The player that starts the game is the one on the left of the big blind. One example of an initial table configuration is on the Figure 1. The dealer is the player F and the small and big blind players are respectably A and B seats.

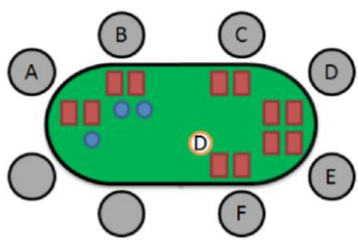

**Fig. 1.** Poker table initial configuration

After configuring the table, the game begins. The game is composed by four rounds (Pre-Flop, Flop, Turn, River) of betting. In each round the player can execute one of the following actions: Call (cover the highest bidder), Raise (exceed the highest bidder), Fold (forfeit the hand).

In any game round, the last player standing wins the game and therefore the pot. If the River round finishes, the player that wins is the one with the highest ranked hand.

### 2.2 Opponent Modeling

Since Texas Hold'em is a game of incomplete information, it is essential to model the opponents in order to predict their actions. By predicting the opponents' actions, the player is able to optimize his profit.

One good example of opponent modeling is the Sklansky groups [3] which define types of players and common actions that they take for each group of cards.

## 3 Related Work

There is some work related to data mining in poker, more particularly the approach of seeing Poker as a classification problem. For instance, in this work [4], the author built a framework that used supervised learning algorithms to copy human tactics in order to create Poker agents. Classification algorithms are also widely used to model opponents. Good examples are [5], where a poker classification system was built that makes decisions based on game observations of hand value, hand risk and player aggressiveness; [6], where were created agents using evolutionary neural networks.

There are also opponent modeling techniques based on simple techniques like measuring the actions of the opponents. Great deal of these techniques are based on real professional poker players' strategies such as David Sklansky, who published one of the most renowned books on Poker strategy [3]. Based on his book, this Ph.D. thesis [7] defined a player classification based on the percentage of hands that a player folds and his aggression factor (Equation 1).

$$AgressionFactor = \frac{NumberBets + NumberRaises}{NumberCalls} \quad (1)$$

If the player folds 72% or more hands then he is considered to be a tight player else he is loose player. Regarding the aggression factor (AF), if the player has an AF above 1 then he is aggressive, else he is considered to be passive. Through this simple classification it is possible to know which hands will be likely played by each opponent.

In these articles [1,2], the concept of Hand Strength and Hand Potential, that was previously defined here [7], was slightly modified to support opponent modeling. Hand strength is the probability that a given hand is better than any other possible hand. Instead of generating all possible two card remaining hands, the authors suggest that one should only generate the hands with cards that the opponents likely have. The same idea could be applied to the Hand Potential algorithm, which is the probability of a given hand improves to be the best at the final round.

Despite all the breakthroughs achieved by known research groups like Computer Poker Research Group[1] and individuals, no artificial poker playing agent is presently known to be capable of beating the best human players.

To the date there are very few approaches that use clustering algorithms to define types of common strategies. This article attempts to fill this gap, by defining new types of strategies which are represented by the frequency of certain actions. The clustering algorithms are used on a very large database in order to extract common playing styles from its players.

### 3.1 Clustering algorithms

Clustering algorithms are unsupervised learning methods that divide a set of objects into several subsets (called clusters) where the objects on the subset are in some way related. Clustering has multiple applications in areas such as machine learning, data mining, imagine analysis and others [8]. A simple 2D axis clustering can be seen on Figure 2.

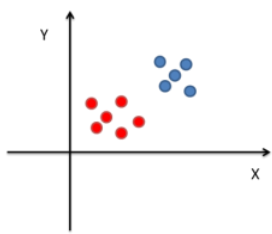

**Fig. 2.** Clustering example

In this figure, a chart is presented with several points. There are clearly two groups of points that can be identified by the proximity between themselves (blue and red groups). This example is very simple, only presenting two dimensions. Clustering can be applied to problems with multiple dimensions that aren't so easily graphically represented.

One particular type of clustering algorithms and the one that is used on this work is partitioning algorithms. This type of algorithms creates various partitions of objects at once and then evaluates them. Some well-known partitioning methods are K-Means or Expectation-maximization (EM).

---

[1] Computer Poker Research Group from University Alberta: http://poker.cs.ualberta.ca/

### 3.2 Clustering in Poker

The publication [9] is a good example of using clustering algorithms to group players with similar playing style, by analyzing their moves. However, the author does not consider important game features like position in table and possible earnings to classify actions, which are considered key aspects of the game strategy, by professional players [3].

Another work about clustering in Poker is [10]. In this work the author uses EM to quickly learn the opponents' strategies using a mixture model of players.

## 4   Work overview

The steps of this work are summarized on Figure 3. We have several sources of Poker Logs that need to be converted to a common format in order to combine information from different sources. Next, a game entity is used to extract information from the logs and thus creating ARFF files. The ARFF files are the files that WEKA [11] uses to describe clusters. The ARFF files in combination with clustering algorithms define game move types for the different game rounds.

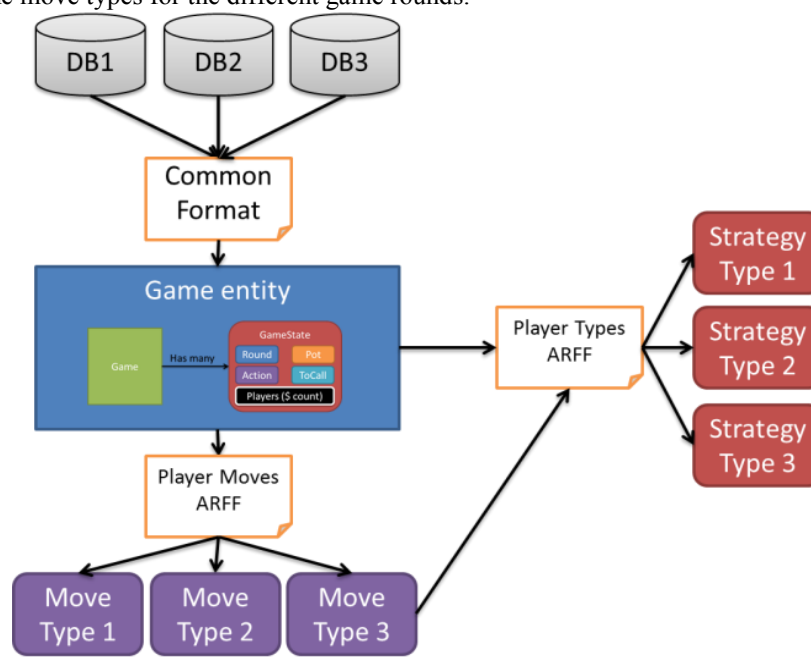

**Fig. 3.** Development steps diagram

By measuring the frequency of each game move by the extracted players, we create a player type ARFF file. This file will be used to define a strategy type, using again clustering algorithms. All these steps will be detailed below on the following sections of the article.

## 5   Poker data extraction

In this work a large database of poker games is needed in order to cover the largest possible number of game situations. There are some desirable requisites that must be accomplished in order to ensure data quality. In this case we need a database that has games between human players and preferably games in real money tournaments. As suggested in [12,13], the money factor is rather important because the players tend to be more careful, because they are risking money, usually using for that reason better tactics.

There are several data sources that might be used to this work. The chosen one was a package of 7.5 GB of casino client game logs[2]. The following table shows some characteristics of the dataset.

**Table 1.** Dataset characteristics.

| Characteristic | Value |
| --- | --- |
| Total Size | 7,51 Gb (zip compressed) |
| Total Games | 51.377.820 |
| Number of Players | 158.035 |
| Total Showdowns | 2.323.538 |
| Showdown Ratio | *4.52%* |

Each file (game log) of the database is composed of a set of Poker games between human players in real money tournaments. Each game contains a set of actions performed by the participant players. One example of game log can be found bellow.

```
Stage #3085270332: Holdem (1 on 1) No Limit $6
Table: LYNCHBURG (Real Money) Seat #6 is the dealer
Seat 6 - nZE2Jjzd6N7Iw/f/mLLEXA ($1,179 in chips)
Seat 4 - PtgusfQqsttogld64pQOGw ($2,214.25 in chips)
nZE2Jjzd6N7Iw/f/mLLEXA - Posts small blind $3
PtgusfQqsttogld64pQOGw - Posts big blind $6
*** POCKET CARDS ***
nZE2Jjzd6N7Iw/f/mLLEXA - Raises $15 to $18
PtgusfQqsttogld64pQOGw - Folds
nZE2Jjzd6N7Iw/f/mLLEXA - returned ($12) : not called
*** SHOW DOWN ***
nZE2Jjzd6N7Iw/f/mLLEXA - Does not show
nZE2Jjzd6N7Iw/f/mLLEXA Collects $12 from main pot
```

---

[2]  The package was found here: http://www.outflopped.com/questions/286/obfuscated-datamined-hand-histories

```
*** SUMMARY ***
Total Pot($12)
Seat 4: PtgusfQqsttogld64pQOGw (big blind) Folded on the POCKET
CARDS
Seat 6: nZE2Jjzd6N7Iw/f/mLLEXA (dealer) (small blind) collected
Total ($12)
```

The information present on game logs is the sequence of actions that were taken by the players on each game. For this reason, to datamine a poker databases a lot of preprocessing is needed, namely extracting information about the state of the game before the action was took. The state of the game can explain why the player decided to play like it did. This way it is possible to know the player's strategy.

To extract the state of the game, first we need to decide which features best represent it. The features are game specific and should represent the reasons that might have led the player to choose certain action. The decided features were:

- **Winning probability**: based on equation 2 from [14]. This formula represents the probability of winning at the Showdown, against a varying number of opponents with the current card hand. The HS is the hand strength, i.e. the probability of the current hand being the best, NPot is the probability of our hand being currently the best but ending up losing and PPot is the probability of our hand not being currently the best but ending up winning the pot.

$$\Pr(\text{win}) = \text{HS} \times (1 - \text{NPot}) + (1 - \text{HS}) \times \text{PPot} \qquad (2)$$

- **Number of opponents**: the number of opponents is a very important feature. It can influence the player's decision because the more opponents you have, the greater the likelihood of one of them having a better hand than the player.
- **Position in table**: the position in table is also important. The last player to "talk" at a Poker table has the advantage of listing all opponents' bets and raises, enabling that player to make a decision based on those actions.
- **Possible earnings**: just how much the player can win with the current pot value. If the pot isn't worth enough the risk, the player could just forfeit it.
- **Minimum bet**: how much money is necessary to continue in the game. Some players continue on the game because the minimum bet is just too low.
- **Game round**: the current game round usually influences the playing style. In later game rounds, the information about the game state is higher as well as the information about the opponents.

Besides the features, some information about the action that was taken from that state is also necessary. This way we can define the player tactic: for a given state of the game, with "X" probability of winning, with "Y" opponents in position "Z", the player took action "A". The player's tactic can be seen as a function that receives the state of the game and returns an action (Equation 3).

$$Tactic(GameState) \rightarrow (Action, Value) \qquad (3)$$

$$GameState = (WinProb., N.Opps., Position, Earnings, Min.Bet)$$

The return value is not only the action type. The value of the bet is also important in No Limit Texas Hold'em, because in this variant the value of the bet can vary from 0 to the player's full stack.

To extract the game state of a database of played poker, an application was built. In this application it was defined the concept of game and game state (Figure 4).

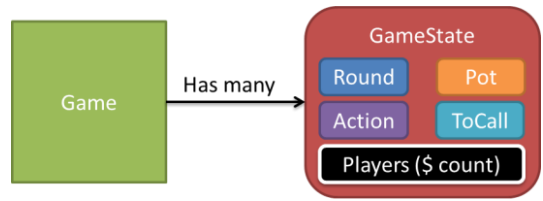

**Fig. 4.** Game and game state definition

The game state has associated the current round, the current pot value, and the current max bet (ToCall). It also has the list of players and the chips that they currently have. The game is composed of various game states that represent each decision of the game made by any player. When a decision is made, the Action parameter of the Game State is set, and the other parameters are updated to create a new game state. The following figure shows an example of creation of a new game state after reading a game log command.

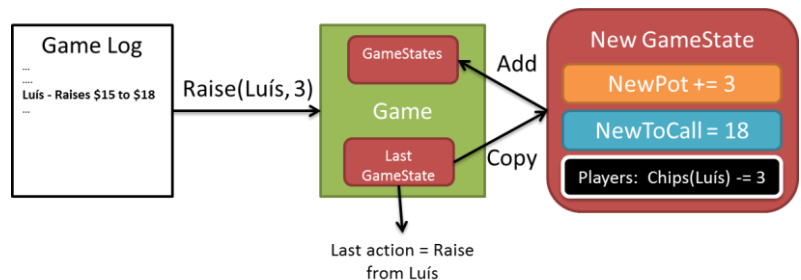

**Fig. 5.** Creation of a new game state after a player raises 3$ on a game log

In this example, the player Luís on the Game Log raised from $15 to 18$. The game entity receives the instruction to create a raise action for player Luís, with the value of 3 chips. First it updates the current last game state action to be the new one (a raise with value 3 from Luís). Next, it copies the state and alters the values of the new state to the updated ones (for instance, now the minimum bet is 18$). Finally, the game entity adds the new state to the end of the game state list.

## 6   Clustering actions in Poker

After extracting game states from the game logs, the clustering algorithms can now be applied. The platform chosen to apply the clustering algorithms on the data was Weka [11]. Weka is a data mining platform that contains plenty clustering algorithms implemented and ready to use.

In order to use Weka, the extracted game states must be converted to an ARFF file, which is the file format that can be interpreted by Weka algorithms. The structure of the extracted ARFF files can be found on the table below (Table 2).

**Table 2.** Actions training set characteristics.

| Name | Type | Description |
| --- | --- | --- |
| win_prob | Numeric {0-100%} | The probability of winning the game |
| position | Nominal {Early, Late} | The position of the player on the table, which defines the order of playing. |
| possible_earnings | Numeric {0-100%} | Minimum amount of money that the player can win by placing this bet. The amount is relative to the current player chips. |
| action | Nominal {Call, Raise} | The action that was took by the player. |
| betted_money | Numeric {0-100%} | The exact quantity of chips that was wagered by the player. The quantity is relative to the current player chips. If the player goes all-in, the value of betted_money is 1. |

As it can be observed, the training features are either numeric or nominal. To be noted that all numerical values were normalized (all of them belong to an interval between 0 and 1). This way it is possible to maintain consistency and comparability between different data instances. One example of generated ARFF file can be found below. Another important note was the removal of minimum bet and number of opponents attributes. The number of opponents was removed because it was redundant since the winning probability formula already considers the number of opponents (the more opponents we have, the less is the probability of winning with the same hand). The minimum bet was removed because it was considered redundant by Weka, by applying the RemoveUseless filter. This filter removes attributes that do not vary much.

```
@relation poker_plays
@attribute win_prob numeric
@attribute position {Early, Late}
@attribute possible_earnings numeric
@attribute action {Call, Raise}
@attribute min_bet numeric
@attribute betted_money numeric
@data
0.358,Late ,0.000,Raise,0.000,0.008
0.434,Late ,0.023,Raise,0.008,0.024
0.412,Early,0.004,Raise,0.004,0.009
0.370,Early,0.027,Call ,0.009,0.009
0.762,Late ,0.000,Raise,0.000,0.007
```

The action types were simplified. There are in total 5 possible types of non-forfeit actions in No Limit Texas Hold'em: Bet, Raise, All-In, Call and Check. Bet is in reality a Raise before anyone betting. All-In is a special Raise where the player puts all his money on the pot. Check is a special Call without putting any money on the table. For this reason, the action attribute was simplified to only have two possible values: Call or Raise. Fold actions are not considered in this problem, because we had no information about the cards of the player when they folded the hands. For this reason, all those hands were discarded.

Two different ARFF files were considered, one for Pre-Flop round and one for Post-Flop rounds (Flop, Turn and River). This was because of the fact that tactics tend to be different in different game rounds. The differences are even greater between rounds before and after the flop, because the quantity of available information is much different. In Pre-Flop there are no community cards, the number of opponents is higher and the number of possible game outcomes is greater.

Some differences between the rounds can be demonstrated in the following graphs representing the distribution of values of the features chosen to represent the types of game movements from a small sample of the data.

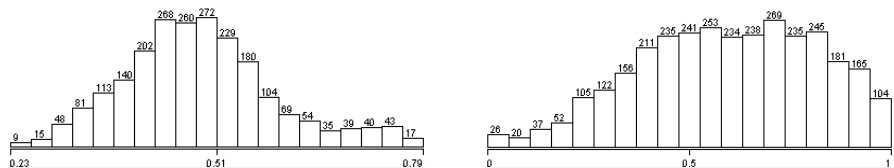

**Fig. 6.** Winning probability distribution (left: Pre-Flop, right: Post-Flop)

In Figure 6 we can compare the distribution of the probability of winning in Pre-Flop and Post-Flop rounds. In Pre-Flop, since there is more hidden information, the players do bets with lower probability of success. In Pre-Flop most of the bets with lower probabilities are checks from the big-blind player, which means that it doesn't pay additional money to see the flop.

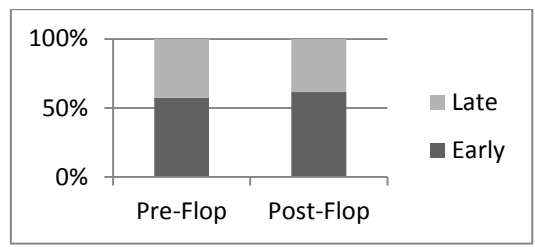

**Fig. 7.** Position distribution in relative frequency

Regarding position, there is no notable difference between the players' behavior in Pre-Flop and Post-Flop (Figure 7). There are more actions in early positions, as was expected, because they are the first players to act.

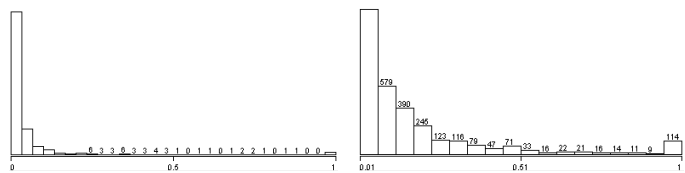

**Fig. 8.** Possible earnings distribution (left: Pre-Flop, right: Post-Flop)

Regarding possible earnings, it is possible to observer that the stakes are much higher after the flop (Figure 8).

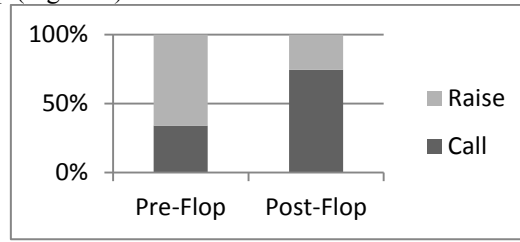

**Fig. 9.** Action type's distribution

Regarding action types it is interesting to note that the trend is reversed completely (Figure 9). In Pre-Flop there are more raise actions than calls, and after the Flop there are much more calls than raises. This can be explained by the fact that there is more money involved in Post-Flop actions which explains the higher frequency of more passive actions, because the players are more afraid of losing the money.

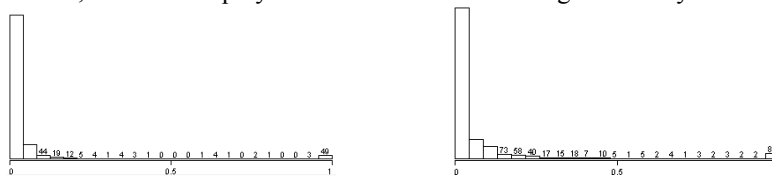

**Fig. 10.** Betted money distribution (left: Pre-Flop, right: Post-Flop)

Finally, regarding to betted money distribution, it is possible to observe that the relative value of the bets slightly increases after the Flop (Figure 10).

The clustering algorithm that was used was Expectation-maximization (EM). The reason for this choice was due to the fact that this algorithm detects the number of clusters, since we do not know how many types of common game movements exist. This algorithm also finds the maximum likelihood which is a popular statistical method to build statistical models from data. So this method can be used to make predictions about new entries weren't used to train the model.

In Pre-Flop round, six different clusters were identified. The centroids of the clusters are as follows.

**Table 3.** Pre-Flop centroids

| Feature | Cluster #0 22% | Cluster #1 9% | Cluster #2 38% | Cluster #3 7% | Cluster #4 9% | Cluster #5 15% |
|---|---|---|---|---|---|---|
| win_prob | 0.4532 | 0.5502 | 0.4893 | 0.5539 | 0.4824 | 0.4703 |
| position | Early | Early/Late | Late | Early/Late | Early | Early |
| possible_earnings | 0.0264 | 0.0323 | 0.0056 | 0.3585 | 0.0862 | 0.004 |
| action | Call | Raise | Raise | Call/Raise | Call | Raise |
| betted_money | 0.0093 | 0.0543 | 0.0128 | 0.4755 | 0.0387 | 0.008 |

One of the most common actions in Pre-Flop is a small raise (1% of the money) in a later position with winning probability about 49%.

For post-flop rounds, the Expectation Maximization algorithm resulted in five different clusters. The centroids of the clusters are as follows.

**Table 4.** Post-Flop centroids

| Feature | Cluster #0 13% | Cluster #1 15% | Cluster #2 14% | Cluster #3 17% | Cluster #4 42% |
|---|---|---|---|---|---|
| win_prob | 0.6946 | 0.5297 | 0.7052 | 0.6463 | 0.5306 |
| position | Late | Late | Early/Late | Early | Early |
| possible_earnings | 0.1118 | 0.0867 | 0.6305 | 0.1133 | 0.1186 |
| action | Raise | Call | Call/Raise | Call/Raise | Call |
| betted_money | 0.0574 | 0.0008 | 0.3586 | 0.0558 | 0 |

The most common actions after the flop (Cluster #4 and 42% of the plays) are Checks (Calls without any money involved).

## 7 Clustering player types

After defining common move types both in Pre-Flop and Post-Flop rounds, clustering algorithms were used to determine common strategies. To do this, all players' actions on the database were considered.

Initially a cluster was assigned to each action. To assign a cluster we calculated the Euclidean distance (Equation 4) between the action and each cluster. The assigned cluster is the one with the lowest distance.

$$d(p, q) = \sqrt{\sum_{i=1}^{n}(q_i - p_i)^2} \qquad (4)$$

Therefore, for each player the frequency of actions of each cluster was calculated. The following table describes the file structure ARFF representative of the players' strategies.

**Table 5.** Training set characteristics to determine player strategies.

| Name | Type | Description |
|---|---|---|
| Name | String | The name of the player. |
| Pre_C0 | Numeric [0,1] | Relative frequency of actions in Pre-Flop of the type of cluster 0. |
| Pre_C1 | Numeric [0,1] | Relative frequency of actions in Pre-Flop of the type of cluster 1. |
| Pre_C2 | Numeric [0,1] | Relative frequency of actions in Pre-Flop of the type of cluster 2. |
| Pre_C3 | Numeric [0,1] | Relative frequency of actions in Pre-Flop of the type of cluster 3. |
| Pre_C4 | Numeric [0,1] | Relative frequency of actions in Pre-Flop of the type of cluster 4. |
| Pre_C5 | Numeric [0,1] | Relative frequency of actions in Pre-Flop of the type of cluster 5. |
| Post_C0 | Numeric [0,1] | Relative frequency of actions in Post-Flop of the type of cluster 0. |
| Post_C1 | Numeric [0,1] | Relative frequency of actions in Post-Flop of the type of cluster 1. |
| Post_C2 | Numeric [0,1] | Relative frequency of actions in Post-Flop of the type of cluster 2. |
| Post_C3 | Numeric [0,1] | Relative frequency of actions in Post-Flop of the type of cluster 3. |
| Post_C4 | Numeric [0,1] | Relative frequency of actions in Post-Flop of the type of cluster 4. |

Since the frequencies are relative, the sum of the frequencies of pre-flop or post-flop has to be 1 (Pre_C0 + Pre_C1 + Pre_C2 + Pre_C3 + Pre_C4 + Pre_C5 = 1 and Post_C0 + Post_C1 + Post_C2 + Post_C3 + Post_C4 = 1).

After applying the Expectation Maximization algorithm on this data, the resulting clusters were the follow.

| Feature | Cluster#0 13% | Cluster#1 3% | Cluster#2 21% | Cluster#3 15% | Cluster#4 33% | Cluster#5 10% | Cluster#6 5% |
|---|---|---|---|---|---|---|---|
| Pre_c0 | 0.3345 | 0.1007 | 0.4626 | 0.0093 | 0.2281 | 0.0019 | 0.0212 |
| Pre_c2 | 0.2167 | 0.1909 | 0 | 0.7364 | 0.3157 | 0.0404 | 0.0104 |
| Pre_c3 | 0.0201 | 0.174 | 0 | 0 | 0.0075 | 0.0094 | 0.0535 |
| Pre_c4 | 0.1971 | 0.1836 | 0.0051 | 0.1504 | 0.1009 | 0.6254 | 0.0261 |
| Pre_c5 | 0.2316 | 0.3508 | 0.5322 | 0.1039 | 0.3477 | 0.3228 | 0.8889 |
| Post_c0 | 0.1091 | 0.2132 | 0 | 0.3059 | 0.0807 | 0.1022 | 0.0666 |
| Post_c1 | 0.2033 | 0.0572 | 0 | 0.4297 | 0.2511 | 0.8755 | 0.0765 |
| Post_c2 | 0.1378 | 0.7045 | 0.0578 | 0.1536 | 0.0172 | 0.0222 | 0.1722 |
| Post_c3 | 0.2674 | 0.0126 | 01564 | 0.0604 | 0.1194 | 0 | 0.2684 |
| Post_c4 | 0.2824 | 0.0125 | 0.7858 | 0.0504 | 0.5315 | 0.0002 | 0.4163 |

Some notes about this results. The name feature was removed because it was just a label; it has no influence on the clustering. Another feature that was removed was Pre_C1 which was removed by the *RemoveUseless* filter from Weka.

7 different tactics were clustered from the database. For instance, for cluster#2 players, about half of their actions in Pre-Flop are of type 0 and the other half of type 5. In Post-Flop most of the actions from these players are of type 4, and some are of type 3.

Using this table it is possible to predict the opponents' behavior. If we store and analyze the opponents' actions during the game, we can determine its tactics and therefore predict possible next moves from those players.

To use this opponent modeling methodology, the agents must save the historical actions of the adversaries throughout the game, saving the features present on Tables 3 and 4 for each action. Using the Euclidian distance formula (equation 4), it is possible to determine the current strategy cluster of the opponents. This way, by knowing the strategy cluster of the opponent, it might be possible to predict some opponents' actions, with the frequency of the action in the cluster being the probability of occurrence of that action.

## 8   Conclusions

The analysis of actions in a Poker game presents a very suitable problem to be solved by clustering algorithms. The definition of groups of actions might help the poker agents to become better players. This is because when the agent is playing against any opponent, it can store all opponents' actions, and that can be used to determine the opponents' strategies. By knowing the opponents' strategies, the agent will probably improve its results in future games. In this article 7 different game strategies were extracted from a Poker database. These strategies can be used to model opponents in future poker artificial agent implementations. The future work in this area should focus on that: integrating this methodology of opponent modeling on Poker agents to check if it improves the agent performance.